%% file: iccv2017_conference.tex
\documentclass[10pt,twocolumn,letterpaper]{article}
\usepackage{tikz}
\usepackage{iccv}

\usepackage{times}
\usepackage{epsfig}
\usepackage{graphicx}
\usepackage{amsmath}
\usepackage{amssymb}

\newcommand{\citep}{\cite}
\usepackage[breaklinks=true,bookmarks=false]{hyperref}
\iccvfinalcopy

\ificcvfinal\fi
\def\iccvPaperID{3037} 

\begin{document}

\title{Scaling the Scattering Transform: Deep Hybrid Networks}

\author{Edouard Oyallon\\
D\'epartement Informatique\\
Ecole Normale Sup\'erieure\\
Paris, France\\
{\tt\small edouard.oyallon@ens.fr}
\and
Eugene Belilovsky\\
University of Paris-Saclay\\ INRIA
and KU Leuven\\
{\tt\small eugene.belilovsky@inria.fr}
\and
Sergey Zagoruyko\\
 Universit\'e Paris-Est\\
  \'Ecole des Ponts ParisTech\\
 Paris, France\\
{\tt\small sergey.zagoruyko@enpc.fr}
}

\maketitle

\input{intro_abstract}

\section{Scattering Networks and Hybrid Architectures}
\label{hyb}




We introduce the scattering transform and motivate its use as a generic input for supervised tasks. A scattering network belongs to the class of CNNs whose filters are fixed as wavelets \citep{oyallon2015deep}. The construction of this network has strong mathematical foundations \citep{mallat2012group}, meaning it is well understood, relies on  few parameters and is stable to a large class of geometric transformations.  In general, the parameters of this representation do not need to be adapted to the bias of the dataset \citep{oyallon2015deep}, making its output a suitable generic representation.
 
We then propose and motivate the use of  supervised CNNs built on top of the scattering network. Finally we propose a supervised encodings of scattering coefficients using 1x1 convolutions, that can retain interpertability and locality properties.



\subsection{Scattering Networks}
\label{scatnet}

\input{scattering_network}

\subsection{Cascading a supervised Deep architecture}
\input{supervision_on_top}

\subsection{Shared Local Encoder for Scattering Representations}
\label{supervised_encoding}
\input{supervised_encoding}

\section{Local Encoding of Scattering}
\label{sle_sec}
We evaluate the supervised SLE on the Imagenet ILSVRC2012 dataset. This is a large and challenging natural color image dataset consisting of $1.2$ million training images and $50,000$ validation images, divided into $1000$ classes. We then show some unique properties of this network and evaluate its features on a separate task.
\subsection{Shared Local Encoder on Imagenet}
\label{sec:sle_exp}
\input{SLE.tex}

\subsection{Interprating SLE's first layer}
\label{sec:sle_first_sec}
\label{interpretation}

\input{interpretating.tex}
\section{Numerical performances of hybrid networks}
\label{small2}
\input{small}

\section{Conclusion}
\input{conclusion}

{\small
\bibliographystyle{ieee}
\bibliography{iccv}
}

\end{document}

%% file: intro_abstract.tex
\begin{abstract}


We use the scattering network as a generic and fixed initialization of the first layers of a supervised hybrid deep network. We show that early layers do not necessarily need to be learned, providing the best results to-date with pre-defined representations while being competitive with Deep CNNs. Using a shallow cascade of $1\times1$ convolutions, which encodes scattering coefficients that correspond to spatial windows of very small sizes, permits to obtain AlexNet accuracy on the imagenet ILSVRC2012. We demonstrate that this local encoding explicitly learns invariance w.r.t. rotations. Combining scattering networks with a modern ResNet, we achieve a single-crop top 5 error of $11.4\%$ on imagenet ILSVRC2012, comparable to the Resnet-18 architecture, while utilizing only 10 layers. We also find that hybrid architectures can yield excellent performance in the small sample regime, exceeding their end-to-end counterparts, through their ability to incorporate geometrical priors. We demonstrate this on subsets of the CIFAR-10 dataset and on the STL-10 dataset. 
\end{abstract}


\section{Introduction}
Image classification is a high dimensional problem that requires building lower dimensional representations that reduce the non-informative images variabilities. For example, some of the main source of variability are often due to geometrical operations such as translations and rotations. An efficient classification pipeline necessarily builds invariants to these variabilities. Deep architectures build representations that lead to state-of-the-art results on image classification tasks \citep{he2015deep}. These architectures are designed as very deep cascades of non-linear end-to-end learned modules \citep{lecun2010convolutional}.  When trained on large-scale datasets they have been shown to produce representations that are transferable to other datasets \citep{zeiler2014visualizing,huh2016makes}, which indicate they have captured generic properties of a supervised task that consequently do not need to be learned. Indeed several works indicate geometrical structures in the filters of the earlier layers \cite{krizhevsky2012imagenet,waldspurger2015these} of Deep CNNs. However, understanding the precise operations performed by those early layers is a complicated \cite{szegedy2013intriguing,oyallon2017building} and possibly intractable task. In this work we investigate if it is possible to replace these early layers, by simpler cascades of non-learned operators that reduce variability while retaining discriminative information.

Indeed, there can be several advantages to incorporating pre-defined geometric priors, via a hybrid approach of combining pre-defined and learned representations. First, end-to-end pipelines can be data hungry and ineffective when the number of samples is low. Secondly, it could permit to obtain more interpertable classification pipelines which are amenable to analysis. Finally, it can reduce the spatial dimensions and the required depth of the learned modules.

A potential candidate for an image representation is the SIFT descriptor \cite{lowe1999object} that was widely used before 2012 as a feature extractor in classification pipelines \cite{sanchez2011high,sanchez2013image}. This representation was typically encoded via an unsupervised Fisher Vector (FV) and fed to a linear SVM. However, several works indicate that this is not a generic enough representation to build further modules on top of  \cite{le2011learning,bo2013multipath}. Indeed end-to-end learned features produce substantially better classification accuracy.
A major improvement over SIFT can be found in the scattering transform \cite{mallat2012group,bruna2013invariant,sifre2013rotation}, which is a type of deep convolutional network, which permits to retain discriminative information normally discarded by methods like SIFT while introducing geometric invariances and stability. Scattering transforms have been shown to already produce representations that lead to the top results on complex image datasets when compared to other unsupervised representations (even learned ones) \citep{oyallon2015deep}. This makes them an excellent candidate for the initial layers of a deep network. We thus investigate the use of scattering as a generic representation to combine with deep neural networks.

Related to our work \cite{perronnin2015fisher} proposed a hybrid representation for large scale image recognition combining a predefined representation and Neural Networks (NN), that uses Fisher Vector encoding of SIFT and leverages NNs as scalable classifiers. In contrast we use the scattering transform in combination with convolutional architectures. Our main contributions are as follows: First, we demonstrate that using supervised local descriptors, obtained by shallow $1\times1$ convolutions, with very small spatial window sizes permits to obtain AlexNet accuracy on the imagenet classification task (Subsection \ref{supervised_encoding}). We show empirically these encoders build explicit invariance to local rotations (Subsection \ref{sec:sle_first_sec}). Second, we propose hybrid networks that combine scattering with modern CNNs  (Section \ref{small2}) and show that using scattering and a ResNet of reduced depth, we obtain similar accuracy to ResNet-18 on Imagenet (Subsection \ref{hybimnet}). Finally, we demonstrate in Subsection \ref{verysmall} that scattering permits a substantial improvement in accuracy in the setting of limited data.  

Our highly efficient GPU implementation of the scattering transform is, to our knowledge, orders of magnitude faster than any other implementations, and allows training very deep networks applying scattering on the fly. Our scattering implementation \footnote{\url{http://github.com/edouardoyallon/pyscatwave}} and pre-trained hybrid models \footnote{\url{http://github.com/edouardoyallon/scalingscattering}}are available.


\if false
Deep architectures build generic and low-dimensional representations that lead to state-of-the-art results on complex tasks such as classification \citep{he2015deep},  game strategies  \citep{silver2016mastering}, or image generation \citep{radford2015unsupervised}.
These architectures are designed as very deep cascades of non-linear end-to-end learned modules. It means that contrary to methods using predefined representations \cite{ref}, they can adapt their representation to a specific task such that they do not reduce important variabilities relatively to their objective. When trained on large-scale datasets they have been shown to produce representations that are transferable to other datasets \citep{zeiler2014visualizing,huh2016makes}, which indicate they have captured generic properties of a supervised task. However, understanding the nature of the internal layers is a difficult task \citep{szegedy2013intriguing}.

Until 2012, unsupervised pipelines composed of predefined features and possibly , 
have led to state of the arts on challenging datasets, including the imagenet. Usually in such an approach, generic local \cite{otero2015anatomy} SIFT decriptors  \cite{lowe1999object} are densely extracted at the initial layer. A major improvement over the SIFT can be found in the Scattering transform \cite{} which permits to retain discriminative information normally discarded. They consist of a cascade of  predefined wavelet transforms and modulus nonlinearities followed by a final spatial averaging and have proven to be successful  in classification tasks involving textures \citep{bruna2013invariant,sifre2013rotation}, hand-written digits \citep{bruna2013invariant}, and sound \citep{anden2014deep}. It is a deep generic interpretable representation that has also shown to lead to the top results on complex image datasets when compared to other unsupervised representations (even learned ones) \citep{oyallon2015deep}. Nevertheless, these representations do not adapt to the specific bias of each dataset and there can be a large performance gap between supervised and unsupervised representations \citep{oyallon2015deep}. 

While several works \cite{???} suggest that cascading a deep networks on top of pre-defined features is not efficient, more recently \cite{perronnin2015fisher} showed it was possible to combine a supervised neural network with Fisher Vectors (FVs) learned from SIFTs, and obtain an accuracy that approaches that of the AlexNet \cite{krizhevsky2012imagenet}. 
Indeed, incorporating pre-defined geometric priors, via a hybrid approach might be very helpful in the case of limited samples (common in application like medical imaging, as few high-dimensional samples are available and pretrained CNN features are unefficient \cite{}. It could also permit to obtain more interpertable classification pipelines which are amenable to analysis, and is likely to reduce the required depth of the learned modules.

In this work, we cascade deep neural network architectures on top of Scattering Networks. Notably, we introduce a \textit{Shared Local Encoder}, that is a supervised encoder of scattering representations which lead to AlexNet accuracy, while being relatively shallow. The learned portion consists of only three $1\times 1$ convolutional layers and 2 fully connected layers. We show it builds explicit invariances to the rotation group. Secondly, we show that it is possible to achieve performance comparable to similar accuracies on ILSVRC2012 by cascading a Resnet \cite{he2015deep,zagoruyko2016wide} on top of a Scattering Network. Finally, we exhibit the state of the art on unsupervised CIFAR10 and on the STL10 datasets, without using the unlabeled samples.

Section \ref{hyb}  describes how the scattering network is constructed, and explains the use of supervision on top of scattering, while the last section motivates the use of a Shared Local Encoder. Section \ref{sle_sec} analyses the performance of our encoder on ILSVRC2012, its first layer and the discriminability properties of the inner layers. Finally, we analyze the performances of our network in Section \ref{small2}, with standard architectures, on CIFAR10, ILSVRC2012 and STL10. Numerical experiments are reproducible; code and models will be available online.
\fi


%% file: scattering_network.tex



\begin{figure}
\begin{center}
\label{archi}

\begin{tikzpicture}
\node at (-1,0) [draw=none,line width=0] (x) {$x$};

\node at (0,0) [minimum width=0.8cm,draw,color=black!60!green] (W1) {$|W_1|$};
\node at (1.5,0) [minimum width=0.8cm,draw,color=red] (W2) {$|W_2|$};
\node at (2.8,0) [minimum width=0.8cm,draw] (A1) {$A_J$};
\node at (3.8,0) [draw=none,line width=0] (S) {$Sx$};

\draw[->,,black] (x) -- (W1);
\draw[->,black] (W1) -- (W2);
\draw[->,black] (W2) -- (A1);
\draw[->,black] (A1) -- (S);
\draw [-<,black] (0.8,0)  -| (0.8,0.5) --(2.8,0.5)|- (2.8,0.32);
\draw [->,black] (-0.6,0) -| (-0.6,-0.5) --(2.8,-0.5)|- (2.8,-0.28);

\end{tikzpicture}

\end{center}
\caption{A scattering network. $A_J$ concatenates the averaged signals.}
\end{figure}
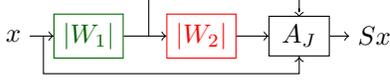

In this section, we recall the definition of the scattering transform. Consider a signal $x(u)$, with $u$ the spatial position index and an integer $J\in \mathbb{N}$, which is the spatial scale of our scattering transform. Let $\phi_J$ be a local averaging filter with a spatial window of scale $2^J$ (here, a Gaussian smoothing function). Applying the local averaging operator, $A_Jx(u)=x\star \phi_J(2^Ju)$ we obtain the zeroth order scattering coefficient, $S_0x(u)=A_Jx(u)$. This operation builds an approximate invariant to translations smaller than $2^J$, but it also results in a loss of high frequencies that are necessary to discriminate signals. 

A solution to avoid the loss of high frequency information is provided by the use of wavelets. A wavelet is an integrable and localized function in the Fourier and space domain, with zero mean. A family of wavelets is obtained by dilating a complex mother wavelet $\psi$ (here, a Morlet wavelet) such that $\psi_{j,\theta}(u)=\frac 1 {2^{2j}}\psi(r_{-\theta}\frac{u}{2^j})$, where $r_{-\theta}$ is the rotation by $-\theta$, and $j\geq 0$ is the scale of the wavelet. A given wavelet $\psi_{j,\theta}$ has thus its energy concentrated at a scale $j$, in the angular sector $\theta$. Let $L\in \mathbb{N}$ be an integer parametrizing a discretization of $[0,2\pi]$. A wavelet transform is the convolution of a signal with the family of wavelets introduced above, with an appropriate downsampling:
\begin{equation*}
W_1x(j_1,\theta_1,u)=\{x\star \psi_{j_1,\theta_1}(2^{j_1}u)\}_{j_1\leq J,\theta_1=2\pi\frac{l}{L},1\leq l\leq L}
\end{equation*}
Observe that $j_1$ and $\theta_1$ have been discretized: the wavelet is chosen to be selective in angle and localized in Fourier. 
With appropriate discretization \cite{oyallon2015deep}, $\{A_Jx,W_1x\}$ is approximatively an isometry on the set of signals with limited bandwidth, and this implies the energy of the signal is preserved. This operator then belongs to the category of multi-resolution analysis operators, each filter being excited by a specific scale and angle, but with the output coefficients not being invariant to translation. To achieve invariance we can not apply $A_J$ to $W_1x$ since it gives a trivial invariant, namely zero.


To tackle this issue, we apply a non-linear point-wise complex modulus to $W_1x$, followed by an averaging $A_J$, which builds a non trivial invariant. Here, the mother wavelet is analytic, thus $|W_1x|$ is  regular \citep{bernstein2013generalized} which implies that the energy in Fourier of $|W_1x|$ is more likely to be contained in a lower frequency domain than $W_1x$. Thus, $A_J$ preserves more energy of $|W_1x|$. It is possible to define $S_1x=A_J|W_1|x$, which can also be written as: $S_1x(j_1,\theta_1,u)=|x\star\psi_{j_1,\theta_1}|\star\phi_J(2^Ju)$; this is the first order scattering coefficients. Again, the use of the averaging  builds an invariant to translation up to $2^J$.

Once more, we apply a second wavelet transform $W_2$, with the same filters as $W_1$, on each channel. This permits the recovery of the high-frequency lost due to the averaging applied to the first order, leading to $S_2x=A_J|W_2||W_1|$, which can also be written as $S_2x(j_1,j_2,\theta_1,\theta_2,u)=|x\star \psi_{j_1,\theta_1}|\star \psi_{j_2,\theta_2}|\star \phi_J(2^Ju)$. We only compute increasing paths, e.g. $j_1< j_2$ because non-increasing paths have been shown to bear no energy \citep{bruna2013invariant}. We do not compute higher order scatterings, because their energy is negligible \citep{bruna2013invariant}. We call $Sx(u)$ the final scattering coefficient corresponding to the concatenation of the order 0, 1 and 2 scattering coefficients, intentionally omitting the path index of each representation. In the case of colored images, we apply independently a scattering transform to each RGB channel of the image, which means  $Sx(u)$ has a size equal to $3\times \big(1+JL+\frac 1 2 J(J-1)L^2\big)$, and the original image is down-sampled by a factor $2^J$ \cite{bruna2013invariant}.

This representation is proved to linearize small deformations \cite{mallat2012group} of images, be non-expansive and almost complete \cite{dokmanic2016inverse,bruna2013audio}, which makes it an ideal input to a deep network algorithm, that can build invariants to this local variability via a first linear operator. We discuss it as an ideal initialization in the next subsection.

%% file: supervision_on_top.tex
We now motivate the use of a supervised architecture on top of a scattering network. 
Scattering transforms have yielded excellent numerical results \cite{bruna2013invariant} on datasets where the variabilities are completely known, such as MNIST or FERET. In these task, the problems encountered are linked to sample and geometric variance and handling these variances leads to solving these problems. However, in classification tasks on more complex image datasets, such variabilities are only partially known as there are also non geometrical intra-class variabilities. Although applying the scattering transform on datasets like CIFAR or Caltech leads to nearly state-of-the-art results in comparison to other unsupervised representations there is a large gap in performance when comparing to supervised representations \cite{oyallon2015deep}. CNNs fill in this gap, thus we consider the use of deep neural networks utilizing generic scattering representations in order to reduce more complex variabilities than geometric ones.



 Recent works \citep{mallat2016understanding,bruna2013learning,jacobsen2017multiscale} have suggested that  deep networks could  build an approximation of the group of symmetries of a classification task and apply transformations along the orbits of this group, like convolutions. This group of symmetry corresponds to some of the non-informative intra class variabilities, which must be reduced by a supervised classifier. \citep{mallat2016understanding} motivates that to each layer corresponds an approximated Lie group of symmetry, and this approximation is progressive, in the sense that the dimension of these groups is increasing with depth. For instance, the main linear Lie group of symmetry of an image is the translation group, $\mathbb{R}^2$. In the case of a wavelet transform obtained by rotation of a mother wavelet, it is possible to recover a new subgroup of symmetry after a modulus non-linearity, the rotation $SO_2$, and the group of symmetry at this layer is the roto-translation group: $\mathbb{R}^2 \ltimes SO_2$. If no non-linearity was applied, a convolution along $\mathbb{R}^2 \ltimes SO_2$ would be equivalent to a spatial convolution. Discovering explicitly the next new and non-geometrical groups of symmetry is however a difficult task \cite{jacobsen2017multiscale}; nonetheless, the roto-translation group seems to be a good initialization for the first layers. In this work, we investigate this hypothesis and avoid learning those well-known symmetries.


Thus, we consider two types of cascaded deep network on top of scattering. The first, referred to as the \textit{Shared Local Encoder} (SLE), learns a supervised local encoding of the scattering coefficients. We motivate and describe the SLE in the next subsection as an intermediate representation between unsupervised local pipelines, widely used in computer vision prior to 2012, and modern supervised deep feature learning approaches. 
The second, referred to as a hybrid CNN, is a cascade of a scattering network and a standard CNN architecture, such as a ResNet \cite{he2015deep}. In the sequel we empirically analyse hybrid CNNs, which permits to greatly reduce the spatial dimensions on which convolutions are learned and can reduce sample complexity.   
  


%% file: supervised_encoding.tex



We now discuss the spatial support of different approaches, in order to motivate our local encoder for scattering. In CNNs constructed for large scale image recognition, the representations at a specific spatial location and depth depend upon large parts of the initial input image and thus mixes global  information. For example, at  depth 2 of \cite{krizhevsky2012imagenet}, the effective spatial support of the corresponding filter is already 32 pixels (out of 224). The specific representations derived from CNNs trained on large scale image recognition are often used as representations in other computer vision tasks or datasets \cite{yosinski2014transferable,zeiler2014visualizing}. 

On the other hand prior to  2012 local encoding methods led to state of the art performance on large scale visual recognition tasks \cite{sanchez2011high}. In these approaches local neighborhoods of an image were encoded using method such as SIFT descriptors \cite{lowe1999object}, HOG \cite{dalal2005histograms}, and wavelet transforms \cite{serre2004realistic}. They were also often combined with an unsupervised encoding, such as sparse coding \cite{boureau2011ask} or Fisher Vectors(FVs) \cite{sanchez2011high}. Indeed, many works in classical image processing or  classification \cite{koenderink1999structure,boureau2011ask,sanchez2011high,perronnin2015fisher} suggests that the local encoding of an image permit to describe efficiently an image. Additionally for some algorithms that rely on local neighbourhoods, the use of local descriptors is essential \cite{lowe1999object}. Observe that a representation based on local non overlapping spatial neighborhood is simpler to analyze, as there is no ad-hoc mixing of spatial information. Nevertheless, on large scale classification, this approach was surpassed by fully supervised learned methods \cite{krizhevsky2012imagenet}.

We show that it is possible to apply, a similarly local, yet supervised encoding algorithm to a scattering transform, as suggested in the conclusion of \cite{perronnin2015fisher}. First observe that at each spatial position $u$, a scattering coefficient $S(u)$ corresponds to a descriptor of a local neighborhood of spatial size $2^J$. As explained in the first Subsection \ref{scatnet}, each of our scattering coefficients are obtained using a stride of $2^J$, which means the final representation can be interpreted as a non-overlapping concatenation of  descriptors. Then, let $f$  be   a cascade of fully connected layers that we identically apply on each $Sx(u)$.  Then $f$ is a cascade of CNN  operators with spatial support size $1\times 1$, thus we write $fSx\triangleq \{f(Sx(u))\}_u$. In the sequel, we do not make any distinction between the $1\times1$ CNN operators and the operator acting on $Sx(u),\forall u$. We refer to $f$ as a \textit{Shared Local Encoder}. We note that similarly to $Sx$, $fSx$ corresponds to non-overlapping encoded descriptors. To learn a supervised classifier on a large scale image recognition task, we cascade fully connected layers on top of the SLE.

Combined with a scattering network, the supervised SLE, has several advantages. Since the input corresponds to scattering coefficients, whose channels are structured, the first layer of $f$ is as well structured. We further explain and investigate this first layer in Subsection \ref{sec:sle_first_sec}. Unlike standard CNNs, there is no linear combinations of spatial neighborhoods of the different feature maps, thus the analysis of this network need only focus on the channel axis.  Observe that if $f$ was fed with raw images, for example in gray scale, it could not build any non-trivial operation except separating different level sets of these images.  

In the next section, we investigate empirically this supervised SLE trained on the ILSVRC2012 dataset.

%% file: SLE.tex
\begin{figure}
\begin{center}
\begin{tikzpicture}

\node at (0,-0.6) (d1) { $\vdots$};
\node at (0,1.8)  (d21) { $\vdots$};
\node at (3.5,-0.6) (d12) { $\vdots$};
\node at (3.5,1.8)  (d22) { $\vdots$};
\node at (0,1.2) [minimum width=0.5cm,line width=0] (S1) {\footnotesize $Sx(u-2^J)$};
\node at (0,0.6) [minimum width=0.5cm,line width=0] (S2) {\footnotesize  $Sx(u)$};
\node at (0,0) [minimum width=0.5cm,line width=0] (S3) {\footnotesize $Sx(u+2^J)$};

\node at (4.5,0.6) [minimum width=0.5cm,line width=0,draw] (F_4) {$F_4$};
\node at (5.5,0.6) [minimum width=0.5cm,line width=0,draw] (F_5) {$F_5$};
\node at (6.5,0.6) [minimum width=0.5cm,line width=0,draw] (F_6) {$F_6$};
\node at (7.2,0.6)  (F_7) {};
\foreach \i in {0,...,2}
{

\foreach \j in {1,...,3}
{

\node at (1*\j+0.5,0.6*\i) [minimum width=0.5cm,line width=0,draw] (\j_F_\i) {$F_\j $};
}
}

\foreach \i in {0,...,2}
{

\foreach \j[evaluate = \j as \jp using int(\j+1)] in {1,...,2} 
{
\draw[->,black] (\j_F_\i) -- (\jp_F_\i);

}
}

\foreach \i in {0,...,2}
{
\draw[->,black] (3_F_\i) -- (F_4);
}

\draw[->,black] (S1) -- (1_F_2);
\draw[->,black] (S2) -- (1_F_1);
\draw[->,black] (S3) -- (1_F_0);

\draw[->,black,dotted] (d22.east) -- (F_4);
\draw[->,black,dotted] (d12.east) -- (F_4);

\draw[->,black] (F_4) -- (F_5);
\draw[->,black] (F_5) -- (F_6);
\draw[->,black] (F_6) -- (F_7);
\end{tikzpicture}
\end{center}
   \caption{Architecture of the SLE, which is a cascade of 3 $1\times 1$ convolutions followed by 3 fully connected layers. The ReLU non-linearity are included inside the $F_i$ blocks for clarity.}
\label{fig:model}
\end{figure}
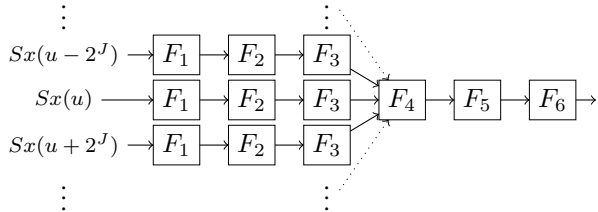

\begin{table}
\begin{center}
\begin{tabular}{|l|c|c|c|}
\hline
\bf Method  &\bf Top 1 &\bf Top 5  \\
\hline


FV + FC    \cite{perronnin2015fisher} &55.6  & 78.4 \\
FV + SVM   \cite{sanchez2011high}          & 54.3 & 74.3\\
AlexNet  & 56.9 &\bf 80.1\\
Scat + SLE  & \bf 57.0&79.6\\
\hline
\end{tabular}
\end{center}
\label{res_1x1}
\caption{Top 1 and Top 5 percentage accuracy reported from one single crop on ILSVRC2012. We compare to other local encoding methods, and SLE outperforms them.  \cite{perronnin2015fisher} single-crop result was provided by private communication.}
\end{table}

We first describe our training pipeline, which is similar to \cite{zagoruyko2016wide}. We trained our network for 90 epochs to minimize the standard cross entropy loss, using SGD with momentum 0.9 and a batch size of 256. We used a weight decay of $1\times10^{-4}$. The initial learning rate is $0.1$, and is dropped off by $0.1$ at epochs $30,50,70,80$. During the training process, each image is randomly rescaled, cropped, and flipped as in \citep{he2015deep}. The final crop size is $224\times 224$. At testing, we rescale the image to a size of 256, and extract a center crop of size $224\times 224$. 

We use an architecture which consists of a cascade of a scattering network, a SLE $f$, followed by fully connected layers. Figure \ref{fig:model} describes our architecture. We select the parameter $J=4$ for our scattering network, which means the output representation has size $\frac{224}{2^4}\times\frac{224}{2^4}=14\times 14$ spatially and 1251 in the channel dimension. $f$ is implemented as 3 layers of 1x1 convolutions $F_1,F_2,F_3$ with layer size 1024. There are 2 fully connected layers of ouput size 1524. For all learned layers we use batch normalization \cite{ioffe2015batch} followed by a ReLU \cite{krizhevsky2012imagenet} non-linearity. We compute the mean and variance of the scattering coefficients on the whole Imagenet, and standardized each spatial scattering coefficients with it.

Table \ref{res_1x1} reports our numerical accuracies obtained with a single crop at testing, compared with local encoding methods, and the AlexNet that was the state-of-the-art approach in 2012. We obtain 20.4\% at Top 5 and 43.0\% Top 1 errors. The performance is analogous to the AlexNet \cite{krizhevsky2012imagenet}. In term of architecture, our hybrid model is analogous, and comparable to that of \cite{sanchez2011high,perronnin2015fisher}, for which SIFT features are extracted followed by FV \cite{sanchez2013image}  encoding. Observe the FV is an unsupervised encoding compared to our supervised encoding. Two approaches are then used: either the  spatial localization is handled either by a Spatial Pyramid Pooling \cite{lazebnik2006beyond}, which is then fed to a linear SVM, either the spatial variables are directly encoded in the FVs, and classified with a stack of four fully connected layers. This last method is a major difference with ours, as the obtained descriptor does not have a spatial indexing  anymore which are instead quantified. Furthermore, in both case, the SIFT are densely extracted which correspond to approximatively $2\,10^4$ descriptors, whereas in our case, only $14^2=196$ scattering coefficients are extracted. Indeed, we tackle the non-linear aliasing (due to the fact the scattering transform is not oversampled) via random cropping during training, allowing to build an invariant to small translations. In Top 1, \cite{sanchez2011high} and \cite{perronnin2015fisher} obtain respectively 44.4\% and 45.7\%. Our method brings a substantial improvement of 1.4\% and 2.7\% respectively.

The BVLC AlexNet \footnote{https://github.com/BVLC/caffe/wiki/Models-accuracy-on-ImageNet-2012-val} obtains a  of 43.1\%  single-crop Top 1  error, which is nearly equivalent to the 43.0\% of our SLE network. The AlexNet has 8 learned layers and as explained before, large receptive fields. On the contrary, our training pipeline consists in 6 learned layers with constant receptive field of size $16 \times 16$, except for the fully connected layers that build a  representation mixing spatial information from different locations. This is a surprising result, as it seems to suggest context information is only necessary at the very last layers, to reach AlexNet accuracy.

We study briefly the local SLE, which has only a spatial extent of $16 \times 16$, as a generic local image descriptor. We use the Caltech-101 benchmark which is a dataset of 9144 image and 102 classes. We followed the standard protocol for evaluation \cite{boureau2011ask} with 10 folds and evaluate per class accuracy, with 30 training samples per class, using a linear SVM used with the SLE descriptors. Applying our raw scattering network leads to an accuracy of $62.8\pm0.7$, and the outputs features from $F_1,F_2,F_3$ brings respectively an absolute improvement of $13.7,17.3,20.1$. The accuracy of the final SLE descriptor is thus  $82.9\pm0.4$, similar to that reported for the final AlexNet final layer in \cite{zeiler2014visualizing} and sparse coding with SIFT \cite{boureau2011ask}. However in both cases spatial variability is removed, either by Spatial Pyramid Pooling \cite{lazebnik2006beyond}, or the cascade of large filters. By contrasts  the concatenation of SLE descriptors are completely local.

%% file: interpretating.tex
\begin{figure}
\begin{center}
\includegraphics[width=1\linewidth]{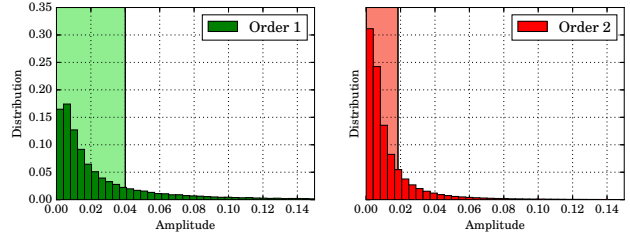} 
\end{center}
   \caption{Histogram of  $\hat F_1$ amplitude for first and second order coefficients. The vertical lines indicate a threshold that is used in Subsection \ref{interpretation} to sparsify $\hat F_1$. Best viewed in color.}
\label{fig:histo}
\end{figure}

\begin{figure}
\begin{center}
\includegraphics[width=1\linewidth]{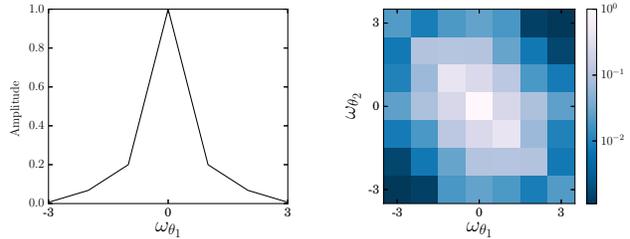} 
\end{center}
   \caption{Energy $\Omega_1\{F\}$  (left)  and $\Omega_2\{F\}$ (right) from Eq. \ref{eq1} for given angular frequencies. Best viewed in color.}
\label{fig:littlehood1}
\end{figure}



Finding structure in the kernel of the layers of depth less than $2$ \cite{waldspurger2015these,zeiler2014visualizing} is  a complex task, and few empirical analyses exist that shed light on the structure \cite{jacobsen2017multiscale} of deeper layers. A scattering transform with scale $J$ can be interpreted as a CNN with depth $J$ \cite{oyallon2015deep}, whose  channels indexes correspond to different scattering frequency indexes, which is a structuration. This structure is consequently inherited by the first layer $F_1$ of our SLE $f$. We analyse $F_1$ and show that it builds explicitly invariance to local rotations, yet also that the Fourier bases associated to rotation are a natural bases of our operator. It is a promising direction to understand the nature of the two next layers.

We first establish some mathematical notions linked to the rotation group that we use in our analysis. For the sake of clarity, we do not consider the roto-translation group.
For a given input image $x$, let $r_\theta . x(u)\triangleq x(r_{-\theta }(u))$ be the image  rotated by angle $\theta$, which corresponds to  the linear action of rotation on images. Observe the scattering representation is  covariant with the rotation in the following sense:
\begin{align*}
S_1(r_\theta.x)(\theta_1,u) &= S_1x(\theta_1-\theta,r_{-\theta}u)\triangleq r_\theta.(S_1x)(\theta_1,u)\\
S_2(r_\theta.x)(\theta_1,\theta_2,u)&=S_2x(\theta_1-\theta,\theta_2-\theta,r_{-\theta}u)\\
&\triangleq r_\theta.(S_2x)(\theta_1,\theta_2,u)
\end{align*}
Besides, in the case of the second order coefficients, $(\theta_1,\theta_2)$ is covariant with rotations, but $\theta_2-\theta_1$ is an invariant to rotation that correspond to a relative rotation. 

Unitary representation framework \cite{sugiura1990unitary} permits the building of a Fourier transform on compact group, like rotations. It is even possible to build a scattering transform on the roto-translation group \cite{sifre2013rotation}. Fourier analysis permits the measurement of the smoothness of the operator and, in the case of CNN operator, it is a natural basis.


We can now numerically analyse the nature of the operations performed along angle variables by the first layer $F_1$ of $f$, with output size $K=1024$. Let us define as $\{F^0_1S_0x,F^1_1S_1x,F^2_1S_2x\}$ the restrictions of $F_1$ to the order 0,1,2 scattering coefficients respectively. Let $1\leq k \leq K$ an index of a feature channel and $1\leq c\leq 3$ the color index. In this case, $F^0_1S_0x$ is simply the weights associated to the smoothing $S_0x$. $F^1_1S_1x$ depends only $(k,c,j_1,\theta_1)$, and $F^2_1$ depends on $(k,c,j_1,j_2,\theta_1,\theta_2)$. We would like to characterize the smoothness of these operators with respect to the variables $(\theta_1,\theta_2)$, because $Sx$ is covariant to rotations.

To this end, we define by $\hat F_1^1$, $\hat F_1^2$ the Fourier transform of these operators along the variables $\theta_1$ and $(\theta_1,\theta_2)$ respectively. These operator are expressed in the tensorial Frequency domain, which corresponds to a change of basis. In this experiment, we normalized each filter of $F$ such that they have a $\ell_2$ norm equal to 1, and each order of the scattering coefficients are normalized as well. Figure \ref{fig:histo} shows the distribution of the amplitude of $\hat F_1^1,\hat F_2^2$. We observe that the distribution is shaped as a Laplace distribution, which is an indicator of sparsity. 

To illustrate that this is a natural basis we explicitly sparsify this operator in its frequency basis and verify that empirically the network accuracy is minimally changed. We do this by thresholding  by $\epsilon$ the coefficients of the operators in the Fourier domain. Specifically we replace the operators $\hat F_1^1$, $\hat F_1^2$ by $1_{|\hat F_1^1|>\epsilon}\hat F_1^1$ and $1_{|\hat F_1^2|>\epsilon}\hat F_1^2$. We select an $\epsilon$ that sets $80\%$ of the coefficients to 0, which is indicated on Figure \ref{fig:histo}. \textit{Without retraining} our network performance degrades by only an absolute value of $2\%$ worse on Top 1 and Top 5 ILSVRC2012. We have thus shown that this basis permits a sparse approximation of the first layer, $F_1$. We now show evidence that this operator builds an explicit invariant to local rotations.

To aid our analysis we introduce the following quantities:
\setlength{\belowdisplayskip}{0pt} \setlength{\belowdisplayshortskip}{0pt}
\setlength{\abovedisplayskip}{0pt} \setlength{\abovedisplayshortskip}{0pt}
\begin{equation}
\Omega_1\{F\}(\omega_1)\triangleq\sum_{k,j_1,c}|\hat{F}^1_1(k,c,j_1,\omega_{\theta_1})|^2
\label{eq1}
\end{equation}

\begin{equation*}
\Omega_2\{F\}(\omega_{\theta_1},\omega_{\theta_2})\triangleq\sum_{k,c,j_1,j_2}|\hat{F}^2_1(k,c,j_1,j_2,\omega_{\theta_1},\omega_{\theta_2})|^2
\end{equation*}
They correspond to the energy propagated by $F_1$ for a given frequency, and permit to quantify the smoothness of our first layer operator w.r.t. the angular variables. Figure \ref{fig:littlehood1} shows variation of  $\Omega_1\{F\}$ and $\Omega_2\{F\}$ along frequencies. For example, if $F_1^1$ and $F_1^2$ were convolutional along $\theta_1$ and $(\theta_1,\theta_2)$, these quantities would correspond to their respective singular values. One sees that the energy is concentrated in the low frequency domain, which indicates that $F_1$ builds explicitly an invariant to local rotations.

%% file: small.tex
We now demonstrate cascading modern CNN architectures on top of the scattering network can produce high performance classification systems. We apply hybrid convolutional networks on the Imagenet ILSVRC 2012 dataset as well as the CIFAR-10 dataset and show that they can achieve performance comparable to modern end-to-end learned approaches. We then evaluate the hybrid networks in the setting of limited data by utilizing a subset of CIFAR-10 as well as the STL-10 dataset and show that we can obtain substantial improvement in performance over analogous end-to-end learned CNNs. 

\subsection{Deep Hybrid CNNs on ILSVRC2012}
\label{hybimnet}
\begin{table}
\begin{center}
\begin{tabular}{|l|c|c|c|c|}
\hline
\bf Method  &\bf Top 1 &\bf Top 5 &\bf Params \\
\hline
AlexNet & 56.9&80.1&61M\\
VGG-16 \cite{han2015learning}&68.5&88.7&138M\\
Scat + Resnet-10 (ours)&68.7&88.6&12.8M\\
Resnet-18 (ours) & 68.9&88.8&11.7M\\
Resnet-200 \cite{zagoruyko2016wide} & \textbf{78.3} & \textbf{94.2} & 64.7M  \\
\hline
\end{tabular}
\end{center}
\caption{ILSVRC-2012 validation accuracy (single crop) of hybrid scattering and 10 layer resnet, a comparable 18 layer resnet, and other well known benchmarks. We obtain comparable performance using analogous amount of parameters while learning parameters at a spatial resolution of 28 $\times$ 28}
\label{tab:imagenet_full}
\end{table}

\begin{table}
\begin{center}
\begin{tabular}{|l|c|c|}
\hline
\bf Method & \bf Accuracy \\
  \hline 

\small \bf Unsupervised Representations   &  \\
Roto-Scat + SVM   \cite{oyallon2015deep}  &82.3\\
ExemplarCNN \cite{dosovitskiy2014discriminative} & 84.3 \\
DCGAN \cite{radford2015unsupervised}& 82.8 \\
Scat + FC   (ours)     & \bf 84.7\\
\hline
\small \bf Supervised and Hybrid    &  \\
Scat + Resnet  (ours) & 93.1       \\
Highway  network \cite{srivastava2015highway}& 92.4 \\
All-CNN \cite{springenberg2014striving}& 92.8 \\
WRN 16 - 8 \cite{zagoruyko2016wide}  & 95.7\\
WRN 28 - 10 \cite{zagoruyko2016wide} &  \textbf{96.0} \\
\hline
\end{tabular}
\end{center}
\label{tab:CIFAR_Main}
\caption{Accuracy of scattering compared to similar architectures on CIFAR10. We set a new state-of-the-art in the unsupervised case and obtain competitive performance with hybrid CNNs in the supervised case.}
\end{table}

We showed in the previous section that a SLE followed by FC layers can produce results comparable with the AlexNet \cite{krizhevsky2012imagenet} on the Imagenet classification task. Here we consider cascading the scattering transform with a modern CNN architecture, such as Resnet \cite{zagoruyko2016wide,he2015deep}. We take the Resnet-18 \cite{zagoruyko2016wide}, as a reference and construct a similar architecture with only 10 layers on top of the scattering network.  We utilize a scattering transform with $J=3$ such that the CNN is learned over a spatial dimension of $28 \times 28$ and a channel dimension of 651 (3 color channels of 217 each). The ResNet-18 typically has 4 residual stages of 2 blocks each which gradually decrease the spatial resolution \cite{zagoruyko2016wide}. Since we utilize the scattering as a first stage we remove two blocks from our model. The network is described in Table \ref{table:arch_imagenet}.

\newcommand{\blocka}[2]{
  \(\left[
      \begin{array}{c}
        \text{3$\times$3, #1}\\[-.1em]
        \text{3$\times$3, #1}
      \end{array}
    \right]\)$\times$#2
}
\newcommand{\blockb}[2]{
 \(\left[
      \begin{array}{c}
        \text{#1}\\[-.1em]
        \text{#1}
      \end{array}
    \right]\)$\times$#2
}
\newcommand{\convsize}[1]{#1$\times$#1}
\newcommand{\convname}[1]{#1}
\def\cellheight{0.34cm}

\begin{table}
  \centering
  \begin{tabular}{|c|c|c|}
    \hline
    Stage & Output size  & Stage details  \\
    \hline
    scattering & $28\times28$ &   $J=3, 651$ channels \\
    \convname{conv1} & \convsize{28} & [256] \\[\cellheight]
    \convname{conv2} & \convsize{28} & \blockb{256}{2}\\[\cellheight]
    \convname{conv3} & \convsize{14} & \blockb{512}{2} \\
    avg-pool & $1\times1$ & [$14\times14$]  \\
    \hline
  \end{tabular}
  \vspace{0.2cm}
  \caption{Structure of Scattering and Resnet-10 used in imagenet experiments. Taking the convention of \cite{zagoruyko2016wide} we describe the convolution size and channels in the Stage details} 
  \label{table:arch_imagenet}
\end{table}

\begin{table}
  \centering
  \begin{tabular}{|c|c|c|}
    \hline
    Stage & Output size & Stage details  \\
    \hline
    scattering & $8\times8$,~$24\times24$ &   $J=2$ \\
    \convname{conv1} & \convsize{8},~\convsize{24} & 16$\times$k , 32$\times$k  \\[\cellheight]
    \convname{conv2} & \convsize{8},~\convsize{24}  &\blockb{32$\times$k}{$n$}\\[\cellheight]
    \convname{conv3} & \convsize{8},~\convsize{12}  &\blockb{64$\times$k}{$n$} \\
    avg-pool & $1\times1$ & [$8\times8$],~[$12\times12$]  \\
    \hline
  \end{tabular}
  \vspace{0.2cm}
  \caption{Structure of Scattering and Wide ResNet hybrid used in small sample experiments. Network width is determined by factor $k$. For sizes and stage details if settings vary we list CIFAR-10 and then the STL-10 network information. All convolutions are of size $3\times3$ and the channel width is shown in brackets for both the network applied to STL-10 and CIFAR-10. For CIFAR-10 we use $n=2$ and for the larger STL-10 we use $n=4$.}
  \label{table:arch_CIFAR}
\end{table}

We use the same optimization and data augmentation procedure described in Section ~\ref{sec:sle_exp} but with learning rate drops at 30, 60, and 80. We find that, when both methods are trained with the same settings of optimization and data augmentation, and when the number of parameters is similar (12.8M versus 11.7 M) the scattering network combined with a resnet can achieve analogous performance (11.4$\%$ Top 5 for our model versus 11.1 $\%$), while utilizing fewer layers. The accuracy is reported in Table \ref{tab:imagenet_full} and compared to other modern CNNs. 

This demonstrates both that the scattering networks does not lose discriminative power and that it can be used to replace early layers of standard CNNs. We also note that learned convolutions occur over a drastically reduced spatial resolution without resorting to pre-trained early layers which can potentially lose discriminative information or become too task specific. 

\subsection{Hybrid Supervised and Unsupervised Representations on CIFAR-10}
\label{cifar10}
We now consider the popular CIFAR-10 dataset consisting of colored images composed of $5\times10^4$ images for training, and $1\times10^4$ images for testing divided into 10 classes. We perform two experiments, the first with a cascade of fully connected layers, that allows us to evaluate the scattering transform as an unsupervised representation. In a second experiment, we again use a hybrid CNN architecture with a ResNet built on top of the  scattering transform. 

For the scattering transform we used $J=2$ which means the output of the scattering stage will be $8\times8$ spatially and 243 in the channel dimension.  We follow the training procedure prescribed in \cite{zagoruyko2016wide} utilizing SGD with momentum of 0.9, batch size of 128, weigh decay of $5\times10^{-4}$, and modest data augmentation of the dataset by using random cropping and flipping. The initial learning rate is 0.1, and we reduce it by a factor of 5 at epochs 60, 120 and 160. The models are trained for 200 epochs in total. We used the same optimization and data augmentation pipeline for training and evaluation in both case. We utilize batch normalization techniques at all layers which lead to a better conditioning of the optimization \citep{ioffe2015batch}. Table \ref{tab:CIFAR_Main} reports the accuracy \textcolor{blue} in the unsupervised and supervised settings and compares them to other approaches.


In the unsupervised comparison we consider the task of classification using only unsupervised features. Combining the scattering transform with a NN classifier consisting of 3 hidden layers, with width $1.1\times10^4$, we show that one can obtain a new state of the art classification for the case of unsupervised features. This approach outperforms all methods utilizing learned and not learned unsupervised features further demonstrating the discriminative power of the scattering network representation. 

In the case of the supervised task we compare to state-of-the-art approaches on CIFAR-10, all based on end-to-end learned CNNs. We use a similar hybrid architecture to the successful wide residual network (WRN) \cite{zagoruyko2016wide}. Specifically we modify the WRN of 16 layers which consists of 4 convolutional stages. Denoting the widening factor, $k$, after the scattering output we use a first stage of $32\times k$. We add intermediate $1\times 1$ to increase the effective depth, without increasing too much the number of parameters. Finally we apply a dropout of 0.2 as specified in \cite{zagoruyko2016wide}.
Using a width of 32 we achieve an accuracy of $93.1\%$. This is superior to several benchmarks but performs worse than the original ResNet \cite{he2015deep} and the wide resnet \cite{zagoruyko2016wide}. We note that training procedures for learning directly from images, including data augmentation and optimization settings, have been heavily optimized for networks trained directly on natural images, while we use them largely out of the box we do believe there are regularization techniques, normalization techniques, and data augmentation techniques which can be designed specifically for the scattering networks.


\subsection{Limited samples setting}
\label{verysmall}
A major application of a hybrid representation is in the setting of limited data. Here the learning algorithm is limited in the variations it can observe or learn from the data, such that introducing a geometric prior can substantially improve performance. We evaluate our algorithm on the limited sample setting using a subset of CIFAR-10 and the STL-10 dataset.

\subsubsection{CIFAR-10}
We take subsets of decreasing size of the CIFAR dataset and train both baseline CNNs and counterparts that utilize the scattering as a first stage. We perform experiments using subsets of 1000, 500, and 100 samples, that are split uniformly amongst the 10 classes. 

We use as a  baseline the Wide ResNet \cite{zagoruyko2016wide} of depth 16 and width 8, which shows near state-of-the-art performance on the full CIFAR-10 task in the supervised setting. This network consists of 4 stages of progressively decreasing spatial resolution detailed in Table 1 of \cite{zagoruyko2016wide}. We construct a comparable hybrid architecture that removes a single stage and all strides, as the scattering already down-sampled the spatial resolution. This architecture is described in Table \ref{table:arch_CIFAR}.  Unlike the baseline, refereed from here-on as WRN 16-8, our architecture has 12 layers and equivalent width, while keeping the spatial resolution constant through all stages prior to the final average pooling.

We use the same training settings for our baseline, WRN 16-8, and our hybrid scattering and WRN-12. The settings are the same as  those described for CIFAR-10 in the previous section with the only difference being that we apply a multiplier to the learning rate schedule and to the maximum number of epochs. The multiplier is set to 10,20,100 for the 1000,500, and 100 sample case respectively. For example the default schedule of 60,120,160 becomes 600,1200,1600 for the case of 1000 samples and a multiplier of 10. Finally in the case of 100 samples we use a batch size of 32 in lieu of 128.

Table \ref{small} corresponds to the averaged accuracy over 5 different subsets, with the corresponding standard error. 
In this small sample setting, a hybrid network outperforms the purely CNN based baseline, particularly when the sample size is smaller. This is not surprising as we incorporate a geometric prior in the representation. 


\begin{table}
\begin{center}

\begin{tabular}{|l|c|c|c|c|}
\hline
\bf Method & \bf 100 & \bf 500 & \bf 1000 \\
  \hline 

WRN 16-8    & 34.7 $\pm$ 0.8  &    46.5 $\pm$1.4& 60.0 $\pm$1.8  \\
Scat + WRN 12-8 & \bf 38.9 $\pm$ 1.2 &\bf 54.7$\pm$0.6 &\bf 62.0$\pm\bf 1.1$ \\

\hline
\end{tabular}
\end{center}
\caption{Mean accuracy of a hybrid scattering in a limited sample situation on CIFAR-10 dataset. We find that including a scattering network is significantly better in the smaller sample regime of 500 and 100 samples.}

\label{small}
\end{table}

\subsubsection{STL-10}
\begin{table}
\begin{center}

\begin{tabular}{|l|c|c|}
\hline
\bf Method & \bf Accuracy \\
\hline

$\substack{\text{\small \bf  Supervised methods}}  $     &  \\
Scat + WRN 19-8 & \bf 76.0 $\pm$ 0.6\\
CNN\cite{swersky2013multi} &  70.1 $\pm$ 0.6\\
\hline
$\substack{\text{\small \bf  Unsupervised methods}}  $     &  \\
Exemplar CNN \cite{dosovitskiy2014discriminative}&  75.4 $\pm$ 0.3\\
Stacked what-where AE \cite{StackedYann} & 74.33 \\
Hierarchical Matching Pursuit (HMP) \cite{bo2013unsupervised}& 64.5$\pm$1\\
Convolutional K-means Network \cite{coates2011selecting} & 60.1$\pm$1\\
\hline
\end{tabular}
\end{center}
\caption{Mean accuracy of a hybrid CNN on the STL-10 dataset. We  find that our model is  better in all cases even compared to those utilizing the large unsupervised part of the dataset.}

\label{tab:small_STL}
\end{table}
The SLT-10 dataset consists of colored images of size $96\times 96$, with only 5000 labeled images in the training set divided equally in 10 classes and 8000 images in the test set. The larger size of the images and the small number of available samples make this a challenging image classification task. The dataset also provides 100 thousand unlabeled images for unsupervised learning. We do not utilize these images in our experiments, yet we find we are able to outperform all methods which learn unsupervised representations using these unlabeled images, obtaining very competitive results on the STL-10 dataset. 

We apply a hybrid convolutional architecture, similar to the one applied in the small sample CIFAR task, adapted to the size of $96\times 96$. The architecture is described in Table \ref{table:arch_CIFAR} and is similar to that used in the CIFAR small sample task. We use the same data augmentation as with the CIFAR datasets. We apply SGD with learning rate 0.1 and learning rate decay of 0.2 applied at epochs 1500,2000,3000,4000. Training is run for 5000 epochs. We use at training and evaluation the standard 10 folds procedure which takes 1000 training images. The averaged result for 10 folds is reported in Table \ref{tab:small_STL}. Unlike other approaches we do not use the 4000 remaining training image to perform hyper-parameter tuning on each fold, as this is not representative of typical small sample situations, instead we train the same settings on each fold. The best reported result in the purely supervised case is a CNN \citep{swersky2013multi,dosovitskiy2014discriminative} whose hyper parameters have been automatically tuned using 4000 images for validation achieving 70.1$\%$ accuracy. The other competitive methods on this dataset utilize the unlabeled data to learn in an unsupervised manner before applying supervised methods. To compare with \cite{hoffer2016deep} we also train on the full training set of 5000 images obtaining an accuracy of $87.6\%$ on the test set, which is substantially higher than $81.3\%$ reported in \cite{hoffer2016deep} using unsupervised learning and the full unlabeled and labeled training set. The competing techniques add several hyper parameters and require an additional engineering process. Applying a hybrid network is on the other hand straightforward and is very competitive with all the existing approaches, without using any unsupervised learning. 

In addition to showing hybrid networks perform well in the small sample regime these results, along with our unsupervised CIFAR-10 results, suggest that completely unsupervised feature learning on natural image data, for downstream discriminative tasks, may still not outperform supervised learning methods and pre-defined representations. One possible explanation is that in the case of natural images, learning in an unsupervised way more complex variabilities than geometric ones ( e.g the rototranslation group), might be very challenging or possibly ill-posed.



%% file: conclusion.tex
This work demonstrates a competitive approach for large scale visual recognition, based on scattering networks, in particular for  ILSVRC2012. When compared with unsupervised representation on CIFAR-10 or small data regimes on CIFAR-10 and STL-10, we demonstrate state-of-the-art results. We build a supervised Shared Local Encoder that permits the scattering networks to surpass other local encoding methods on ILSVRC2012. This network of just 3 learned layers permits analysis on the operation performed.

Our work also suggests that pre-defined features are still of interest and can provide enlightenment on deep learning techniques and to allow them to be more interpretable. Combined with appropriate learning methods, they could permit having more theoretical guarantees that are necessary to engineer better deep models and stable representations.

\subsubsection*{Acknowledgments}
The authors would like to thank  Mathieu Andreux, Matthew Blaschko, Carmine Cella, Bogdan Cirstea, Michael Eickenberg, St\'ephane Mallat for helpful discussions and support. The authors would also like to thank Rafael Marini and Nikos Paragios for use of computing resources. We would like to thank Florent Perronnin for providing important details of their work. This   work   is   funded   by   the   ERC   grant   InvariantClass   320959,   via   a   grant   for   PhD   Students of the Conseil r\'egional d'Ile-de-France (RDM-IdF), Internal Funds KU Leuven, FP7-MC-CIG 334380, an Amazon Academic Research Award, and DIGITEO 2013-0788D - SOPRANO. 